\documentclass[article]{jss}

\usepackage{amsmath}
\usepackage{algorithm}
\usepackage[noend]{algpseudocode}
\usepackage{epstopdf}
\usepackage{float}
\usepackage{breakcites}

\author{
		Vincenzo Lagani\\University of Crete\And\And
        Giorgos Athineou\\University of Crete\And\And
        Alessio Farcomeni\\Sapienza - University of Rome\AND
        Michail Tsagris\\University of Crete\And\And
        Ioannis Tsamardinos\\University of Crete
}

\title{Feature Selection with the \proglang{R} Package \pkg{MXM}: Discovering Statistically-Equivalent Feature Subsets}

\Plainauthor{Vincenzo Lagani, Giorgos Athineou, Alessio Farcomeni, Michail Tsagris, Ioannis Tsamardinos} 
\Plaintitle{Feature Selection with the R Package MXM: Discovering Multiple, Statistically-Equivalent, Predictive Feature Subsets} 
\Shorttitle{Feature Selection with the \proglang{R} Package \pkg{MXM}} 
\Abstract{
The statistically equivalent signature (SES) algorithm is a method for feature selection inspired by the principles of constrained-based learning of Bayesian Networks. Most of the currently available feature-selection methods return only a single subset of features, supposedly the one with the highest predictive power. We argue that in several domains multiple subsets can achieve close to maximal predictive accuracy, and that arbitrarily providing only one has several drawbacks. The SES method attempts to identify multiple, predictive feature subsets whose performances are statistically equivalent. Under that respect SES subsumes and extends previous feature selection algorithms, like the max-min parent children algorithm.\\
SES is implemented in an homonym function included in the \proglang{R} package \pkg{MXM}, standing for mens ex machina, meaning 'mind from the machine' in Latin. 
The \pkg{MXM} implementation of SES handles several data-analysis tasks, namely classification, regression and survival analysis. In this paper we present the SES algorithm, its implementation, and provide examples of use of the \code{SES} function in \proglang{R}. Furthermore, we analyze three publicly available data sets to illustrate the equivalence of the signatures retrieved by SES and to contrast SES against the state-of-the-art feature selection method LASSO. Our results provide initial evidence that the two methods perform comparably well in terms of predictive accuracy and that multiple, equally predictive signatures are actually present in real world data.
}

\Address{
  Ioannis Tsamardinos\\
  Computer Science Department\\
  University of Crete\\
  Voutes Campus\\
  GR-70013 Heraklion, Crete, Greece\\
  Telephone: +30 2810 39 3575\\
  Fax: +30 2810 39 1428\\
  E-mail: \email{tsamard@csd.uoc.gr}\\
  URL: \url{http://www.mensxmachina.org/}
}

\begin{document}


\section[Introduction]{Introduction}
Feature selection is one of the fundamental tasks in the area of machine learning. 
Generally speaking, the process of feature or variable selection aims to identify a subset of features that are relevant with respect to a given task; for example, in regression and classification it is often desirable to select and retain only the subset of variables with the highest predictive power. The main goals of feature selection usually are (a) to improve the performance of a predictive model, (b) to avoid the cost associated with measuring all the features and (c) to provide a better understanding of the predictive model, and by extension of the data, by eliminating useless or redundant features \citep{Guyon2003}. To date, almost all feature selection algorithms return a single feature subset.

\emph{In our experience, it is often the case that multiple feature subsets are approximately equally predictive for a given task}. Low statistical power due to an insufficient sample size can simply make it impossible to distinguish the predictive performance of two or more signatures in a statistically meaningful way. More intriguingly, the physical process that generates the data could be possibly characterized by a high level of redundancy: several of its components can have similar or identical behavior/scope. 
Measurements taken over redundant components would be equivalent to each other, and there would be no particular reason for preferring one over the other for inclusion in a predictive subset. This problem is particularly relevant in biology, where nature uses redundancy for ensuring resilience to shocks or adverse events.

Discovering multiple and statistically equivalent feature subsets has several advantages in our opinion. First, knowing that multiple equally-predictive subsets actually exist increases the understanding of the specific problem at hand. In contrast, identifying a single subset of relevant features can lead to ignore factors that may play an important role for understanding the dynamics of the problem under study. On more practical terms, equally-predictive subsets may differ in terms of the cost/effort needed for measuring their respective components. Thus, providing multiple, alternative subsets can have a great impact in contexts where some factors may be technically difficult or excessively expensive to measure.\\
Recently, algorithms that generate multiple, equivalent feature sets have been developed \citep{Statnikov2010a,Huang2015}, including the Statistically Equivalent Signatures (SES) method \citep{Tsamardinos12}, which is implemented in the \proglang{R} \citep{RCoreTeam2015} \pkg{MXM} package. SES is a constraint-based, feature selection algorithm that attempts to identify multiple, equally-predictive \emph{signatures}, where for signatures we indicate minimal-size sets of features with maximal predictive power. SES subsumes and extends previous work on future selection, particularly the max-min parent children (MMPC) algorithm \citep{Tsamardinos2006a} and related extensions \citep{Lagani2010b,Lagani2013}, by implementing a heuristic method for identifying equivalences among predictors.\\
Other statistical approaches producing several models for the same task exist, for example model averaging \citep{Buckland1997}. In this approach several competitive models are first generated and then combined together for producing a single, global model. The key difference with SES is that model-averaging models can have different predictive capabilities, while SES-retrieved signatures are assumed to be equally predictive. Model averaging methods are already available in several \proglang{R} packages, like \pkg{MuMIn}   \citep{MuMIn_ref}, \pkg{glmulti} \citep{glmulti_ref}, \pkg{AICcmodavg} \citep{AICcmodavg_ref}, \pkg{BMA} \citep{BMA_ref}.\\
Finally, to the best of our knowledge, the \pkg{MXM} package is one of the few open-source code providing implementations of constraint-based feature selection algorithms. The MMPC algorithm has been previously implemented in the \pkg{bnlearn} package \citep{bnlearn_ref} along with several Bayes Network learning methods, and the TETRAD software \citep{Landsheer2010} provides implementations of numerous casual-discovery oriented constraint-based methods. The \proglang{MATLAB} library Causal Explorer \citep{Aliferis2003a} has been the first software offering feature-selection oriented constraint-based methods, but the code is not open-source.\\
In the rest of the paper we present the SES algorithm and detail its characteristics. Moreover, we introduce the \pkg{MXM} package and provide some practical examples for illustrating its use. Finally, we empirically evaluate the results of the SES algorithm on three different data sets, and we contrast our results against the widely used LASSO selection algorithm \citep{glmnet_ref}. Our results support claims that SES is able to return signatures that are statistically equivalent, and whose predictive performances are comparable with the ones of a state-of-the-art feature selection method.\\

\section[Multiple signature selection with SES algorithm]{Multiple signature selection with SES algorithm}
The SES algorithm \citep{Tsamardinos12} belongs to the class of constraint-based, feature selection algorithms \citep{Tsamardinos2006a}, a class of algorithms that  ground their root in the theory of Causal Analysis \citep{spirtes2000}. Principles borrowed from this theory allow for an important result: under some broadly-accepted assumptions, the optimal set of predictors for a given target variable consists in the Markov Blanket (MB) of the variable in the Bayesian Network (BN) representing the data distribution at hand \citep{tsamardinos2003c}. Bayesian Networks \citep{Neapolitan2003} are graphical models that allow compact representations of multivariate distributions under the form of a Direct Acyclic Graph (DAG) and an appropriate parameterization. Nodes in the DAG represent random variables, while edges represent conditional associations. When two nodes are directly connected by an edge, then the association between the two corresponding variables holds in the context of all other variables. Node $A$ is a parent for node $B$ (and $B$ is a child of $A$) if an edge from $A$ is incident to $B$. The MB of a given target $T$ in composed by the set of Parent and Children (PC) of $T$ plus any additional parent of $T$ children (spouses). MMPC was one of the first feature selection methods specifically designed in order to identify the PC set of a given variable. It is interesting to note that PC and BN predictive capabilities are often equivalent in practical applications, while PC is easier to identify \citep{Tsamardinos2006a}. Finally, constraint-based algorithms have recently proven to be able to retrieve highly predictive signatures \citep{Aliferis2010a}.\\
From an algorithmic point of view, given a data set $D$ defined over a set of $n$ variables / predictors $V$ and a target variable $T$ (a.k.a. outcome), constraint-based feature selection methods repetitively apply a statistical test of conditional independence in order to identify the subset of variables that can not be made independent by the outcome given any other subset of variables in $V$. We denote with $ind(X,T|W)$ any statistical test able to provide a $p$~value $p_{XT.W}$ for assessing the null hypothesis that the variables $X$ and $T$ are conditionally independent given a set of variables $W$. Depending on the nature of the variables involved in the test (e.g., categorical, continuous, censored) the most appropriate conditional independence test must be chosen (see Section \ref{condIndSection} for further discussion). Finally, it is worthwhile to note that under some additional assumptions, constraint-based methods have the interesting property of uncovering (part of) the causal mechanism that produced the data at hand.\\
The SES algorithm also retrieves the PC set of target variables, and it subsumes the MMPC algorithm by implementing an additional heuristic in order to retrieve multiple subsets that are possible PC candidates. The iTIE$^*$ algorithm \citep{Statnikov2010a} is based on similar principles, but it adopts a different heuristic for identifying equivalence of features.
SES is summarized in Algorithm \ref{algorithm1} through pseudo code. The proposed method accepts as input a data set $D$, a target variable $T$, a two hyper-parameters\footnote{we define as ``hyper-parameter'' any parameter that is not estimated from the data but that must be provided a priori from the user.}: The threshold $a$ for assessing conditional independence and an upper bound $k$ on the number of variables that can be included in any conditional set. This latter parameter limits the complexity and the computational requirements of the algorithm. The output of the method consists in a set $E$ of variables sets (queues) $Q_i$, $i=1 \ldots n$, such that each queue $Q_i$ contains variables that are 'equivalent' to each other.\\
At initialization, an empty set $S$ of selected variables is created, all variables $V$ are considered for inclusion in $S$ ($R \gets V$, where $R$ is the set of variable considered for inclusion) and each variable is considered equivalent only to itself ($Q_{i} \gets i$). During the main loop the algorithm alternatively attempts to (a) include in $S$ the variable maximally associated with $T$ conditioned on any possible subset of the variables already selected and (b) exclude from $S$ any variable $X$ that is not any more associated with $T$ given any subset $Z$ of other variables in $S$. Once a variable $X$ is excluded from $S$, it cannot be inserted any more.\\
However, before eliminating $X$ from $S$, the algorithm tries to identify any variable $Y$ in $Z$ that is equivalent to $X$, by verifying whether $p_{YT.Z'} > a$ when $Z' \gets ( Z \cup \{X\} ) \setminus \{Y\}$. If such a variable exists, the list of $X$-equivalent variables $Q_{X}$ is added to $Q_{Y}$ (in contrast, the iTIE$^*$ algorithm tests whether $p_{ZT.Y} > a$, i.e., it checks if the whole set $Z$ is equivalent to $Y$). Finally, all equivalence lists $Q_{i}$, $i \in S$, are returned as output.\\
One can then build a predictive signature by choosing \emph{one and only one variable} from each variable set $Q_i$. To fix the ideas, let's assume that $\textbf{E}$ contains three queues, namely $Q_1 = \{X_1, X_4\}$, $Q_3 = \{X_3\}$ and $Q_7 = \{X_7, X_2\}$. Then there are a total of $2 \cdot 1 \cdot 2 = 4$ possible signatures, i.e., $S_a = \{X_1,X_3,X_7\}$, $S_b = \{X_1,X_3,X_2\}$, $S_c = \{X_4,X_3,X_7\}$, $S_d = \{X_4,X_3,X_2\}$. In contrast, the sets $\{X_1, X_2\}$ and $\{X_1, X_4, X_3, X_7\}$ are not valid equivalent signatures, since the first does not select any variable from $Q_3$ and the latter includes two variables from the same queue ($Q_1$).

\makeatletter
\def\BState{\State\hskip-\ALG@thistlm}
\makeatother
\begin{algorithm}
\centering
\begin{algorithmic}[1]
\BState \textbf{Input}:\\
\text{Data set on $n$ predictive variables $V$}\\
\text{Target variable $T$}\\
\text{Significance threshold $a$}\\
\text{Max conditioning set $k$}\\

\BState \textbf{Output}:\\
\text{A set $E$ of size n of variables sets $Q_{i}$, $i=1, \ldots, n$ such that one can construct}\\ {a signature by selecting one and only one variable from each set $Q_{i}$}
\\\\
\text{//Remaining variables}
\State $R \gets V$\\
\text{//Currently selected variables}
\State $S \gets \oslash$\\
\text{//Sets of equivalences}
\State $ Q_{i} \gets i$ {, for $i = 1, \ldots, n$}\\

\While{$R \ne \oslash$}
    \For{all $X \epsilon \{R \cup S \}$}
        \If{$ \exists Z \subseteq S \setminus \{X\}, \vert Z \vert \le k, s.t., p_{XT.Z} > a $}
            \State{$R \gets R \setminus \{X\}$}
            \State{$S \gets S \setminus \{X\}$}
            \\
            \State{\text{//Identify statistical equivalences, i.e., $X$ and $Y$ seem interchangeable}}
            \If{$ \exists Y \epsilon Z, s.t., Z' \gets ( Z \cup \{X\} ) \setminus \{Y\}, p_{YT.Z'} > a$}
                \State{$Q_{Y} \gets Q_{Y} \cup Q_{X}$}
            \\\EndIf{\textbf{endif}}
        \\\EndIf{\textbf{endif}}
    \\\EndFor{\textbf{endfor}}
    \\
    \State $M = argmax_{\{X \epsilon R\}} min_{\{Z \subseteq S , [Z] \le k\}}-p_{XT.Z}$
    \State{$R \gets R \setminus \{M\} $}
    \State{$S \gets S \cup \{M\}$}
    \\
\EndWhile{\textbf{endwhile}}
\\
\BState{Repeat the for-loop one last time}\\
\text{//Pack all the identified equivalences in one data structure}
\State{$E \gets \oslash$}
\For{all $i \epsilon S$}
    \State{$E \gets E \cup \{Q_{i}\} $}\\
\EndFor{\textbf{endfor}}\\\\
\Return{$E$}

\end{algorithmic}
\caption{SES}\label{algorithm1}
\end{algorithm}

\section[Package implementation]{Package implementation}
The  \pkg{MXM} package for \proglang{R} currently implements the algorithm SES along with a variety of different conditional independence test. Conditional independence tests $ind(X,T|W)$ are the cornerstones upon which constraint-based algorithms, including SES, are built. Interestingly, constraint-based algorithms can be applied to different types of data as far as they are equipped with a conditional independence test suitable for the data at hand. Similarly, the \code{SES} function can be applied on different types of outcome (continuous, time-to-event, categorical) and predictors (continuous, categorical, mixed) if the appropriate test is provided. We have implemented a number of these tests in order to grant the user a wide flexibility in terms of the data analysis tasks that can be addressed with the \pkg{MXM} package. The \code{SES} function even allows the users to provide their custom function for assessing conditional independence. The following subsections further illustrate and elaborate upon the implemented functions.\\

\subsection[Conditional independence tests]{Conditional independence tests}
\label{condIndSection}
Assessing dependence among two random variables is one of the oldest problems in statistics, yet it is far from being solved \citep{Reshef2011DetectingSets, Simon2014Comment2011}. Evaluating the conditional independence $ind(X,T|W)$ is further complicated by the presence of the conditioning set $W$. Moreover, one may desire to deal with cases when $X$, $T$ and $W$ are all continuous, categorical, or mixed. Two methods often used in the area of constraint-based algorithms are the Fisher's \citep{Fisher1923} and the $G^2$ tests \citep{spirtes2000}. The former is based on partial correlations and assumes continuous measurements and multivariate normality, while the latter is based on contingency tables and can be applied on categorical data. Both tests are implemented in \pkg{MXM} in the functions \code{testIndFisher} and \code{gSquare}, respectively.\\
Beside these two functions, we have devised and implemented a number of different conditional independence tests following an approach presented by \cite{Lagani2010b}. Briefly, $ind(X,T|W)$ can be assessed by comparing two nested-models, $mod_0$ and $mod$, obtained by regressing the target variable $T$ on the conditioning set $W$ alone and on the conditioning set along with the candidate variable $X$, respectively. In \proglang{R} language formulas, $mod_0 = T \sim W$ and $mod = T \sim X + W$. The $p$~value $p_{XT.W}$ can be computed through a log-likelihood ratio or $\chi^2$ test, depending on the nature of the two models. Table \ref{table:CondIndTests} summarizes the conditional independence tests implemented in \pkg{MXM}. Each test is characterized by (a) the type of outcome and predictors it can deal with and (b) the regression method used \cite[if the test is derived according to the approach from ][]{Lagani2010b}. Some of the tests have the option of employing a robust version of the original regression method.

\begin{table}[!ht]
\centering
\begin{tabular}{|p{3.1cm}| p{3cm} | p{2cm}| p{3.5cm}| p{1.5cm}|}
	\hline \rule[-2ex]{0pt}{5.5ex} Name & Outcome & Predictors & Regression & Robust option \\
	\hline \rule[-2ex]{-2pt}{5.5ex} \code{testIndFisher} & Continuous & Continuous & Linear regression & Yes\\
  	\hline \rule[-2ex]{-2pt}{5.5ex} \code{testIndSpearman} & Continuous & Continuous & Linear regression & No\\
	\hline \rule[-2ex]{-2pt}{5.5ex} \code{gSquare} & Categorical & Categorical & Contingency tables & No\\
	\hline \rule[-2ex]{-2pt}{5.5ex} \code{testIndReg} & Continuous & Mixed & Linear regression & Yes \\
	\hline \rule[-2ex]{-2pt}{5.5ex} \code{testIndRQ} & Continuous & Mixed & Quantile regression & No \\
	\hline \rule[-2ex]{-2pt}{5.5ex} \code{testIndBeta} & Proportions & Mixed & Beta regression & No \\
	\hline \rule[-2ex]{-2pt}{5.5ex} \code{testIndPois} & Count variable & Mixed & Poisson regression & No \\
	\hline \rule[-2ex]{-2pt}{5.5ex} \code{testIndNB} & Overdispersed count variable & Mixed & Negative binomial regression & No \\
   	\hline \rule[-2ex]{-2pt}{5.5ex} \code{testIndZIP} & Zero inflated count data & Mixed & Zero inflated poisson regression & No \\
	\hline \rule[-2ex]{-2pt}{5.5ex} \code{censIndCR} & Survival outcome & Mixed & Cox regression & No \\
   	\hline \rule[-2ex]{-2pt}{5.5ex} \code{censIndWR} & Survival outcome & Mixed & Weibull regression & No \\
     \hline \rule[-2ex]{-2pt}{5.5ex} \code{testIndClogit} & Case-control & Mixed & Conditional logistic regression & No \\
    \hline \rule[-2ex]{-2pt}{5.5ex} \code{testIndLogistic} & Categorical & Mixed & Logistic regression & No \\
    \hline \rule[-2ex]{-2pt}{5.5ex} \code{testIndLogistic} & Categorical & Mixed & Logistic regression & No \\
    \hline \rule[-2ex]{-2pt}{5.5ex} \code{testIndSpeedglm} & Continuous, binary or counts & Mixed & Linear, logistic and poisson regression & No \\
    \hline
\end{tabular}
\caption{Conditional independence tests implemented in \pkg{MXM}. For each test the type of outcome, predictors, and regression method is specified in the respective columns. Some of the tests can also employ a robust version of their respective regression method.}
\label{table:CondIndTests}
\end{table}

\subsection[SES implementation]{SES implementation}
The \code{SES} function has been implemented with the aim of making its usage as intuitive as possible for the user. Only two inputs are required, the matrix of predictor variables \code{dataset} and the outcome variable \code{target}. The first can be either a numeric matrix, a data frame or an object of the class \code{ExpressionSet} from the Bioconductor package \pkg{affy} \citep{affy_ref}. The outcome can be encoded either as a numerical vector, a (ordered) factor, or an object of the \code{Surv} class defined in package \pkg{survival} \citep{survival_ref}.\\
Depending on the \code{dataset} and \code{target} specified by the user, the \code{SES} function is able to automatically select the data analysis task to perform and the conditional independence test to use:

\begin{enumerate}

\item Binary classification: in a binary classification task the objective of the analysis is to find the model that better discriminates between two classes. An example of binary classification is discerning among Alzheimer and healthy patients on the basis of clinical data. If the \code{target} variable is a factor with two levels, the \code{SES} function automatically assumes that the problem is a binary classification task. The default conditional independence test used is \code{testIndLogistic}.

\item Multi-class classification: this tasks is similar to the binary classification task, but more than two classes are present. These classes may have an intrinsic order, e.g., they represent progressively more severe stages of the same cancer, or they may be independent by each other, as for totally different types of diseases. In the first case an ordered factor should be provided as \code{target} variable, while a non-ordered factor should be provided in the second case. In both cases the default conditional independence test is \code{testIndLogistic}, which automatically switches between multinomial logistic (nominal outcome) or ordered logit (ordinal outcome) regression \citep{Lagani2013}.

\item Regression: In this case the scope of the analysis is to predict the values of a continuous target, for example the expression of a given gene. For regression tasks the \code{target} variable should be encoded as a numeric vector, and depending whether \code{dataset} contains solely continuous or mixed (categorical/continuous) predictors the \code{SES} function uses the \code{testIndFisher} or the \code{testIndReg} as conditional independence test, respectively.

\item Time-to-event / Survival analysis: the scope of this type of analysis is to estimate the incidence of an event over time. Survival analysis is conceptually similar to regression, but differs for the presence of censorship, i.e., the exact time-to-event may be unknown for part of the samples. Time-to-event analysis requires a \code{Surv} object (package \pkg{survival}) as \code{target} variable, and the default conditional independence test is the \code{testIndCR}.

\end{enumerate}

The user can override the default behavior of the \code{SES} function by directly specifying a test to use or by providing a custom function for assessing conditional independence. For example, the user can decide to use the \code{testIndPois} instead of the \code{testIndFisher} if \code{target} contains count values. The user can furthermore control the operation of the SES algorithm by specifying the values for the hyper-parameters $a$ and $k$. The signature of the method along with a short explanation of its arguments now follows:

\begin{CodeInput}
R> SES(target, dataset, max_k = 3, threshold = 0.05, test = NULL, 
+    user_test = NULL, hash = FALSE, hashObject = NULL, robust = FALSE, 
+    ncores = 1)
\end{CodeInput}

\begin{itemize}
\item \code{target}: the class variable, encoded as a vector, factor, an ordered factor or a Surv object. If a character or an integer is provided, then the corresponding column in \code{dataset} is used as target.
\item \code{dataset}: either a data frame or a matrix (columns = variables , rows = samples). Alternatively, an \code{ExpressionSet} object from the package \pkg{BioBase} \citep{biobase_ref}.
\item \code{max_k}: the maximum size for the conditioning set to use in the conditional indepedence test.
\item \code{threshold}: cut-off value for assessing $p$~values significance.
\item \code{test}: the conditional independence test to use. If \code{NULL}, the \code{SES} function automatically determines a suitable test depending on \code{target} and \code{dataset}.
\item \code{user_test}: a user-defined conditional independence test (provided as a closure type object). If \code{user_test} is provided, the \code{test} argument is ignored.
\item \code{hash}: logical variable which indicates whether to store (\code{TRUE}) or not (\code{FALSE}) the statistics calculated during \code{SES} execution in a hash-type object. Default value is \code{FALSE}. If \code{TRUE} the hash Object is produced and returned in the \code{SES} output.
\item \code{hashObject}: a list with the hash objects generated in a previous run of \code{SES}. Each time \code{SES} runs with \code{hash=TRUE} it produces a list of hash objects that can be re-used in order to speed up next runs of \code{SES}.
\item \code{robust}: A boolean variable which indicates whether (\code{TRUE}) or not (\code{FALSE}) to use a robust version of the statistical test if it is available. It takes more time than a non-robust version but it is suggested in case of outliers. Default value is \code{FALSE}.
\item \code{ncores}: An integer value indicating the number of CPUs to be used in parallel during the first step of the SES algorithm, where univariate associations are examined.
\end{itemize}

Internally, the \code{SES} function has been optimized in order to improve computational performances. Particularly, the code has been optimized at three different levels:
\begin{itemize}
\item Algorithmic level: constraint-based algorithms\textquotesingle ~computational time is mainly spent for assessing conditional independence. We adopted an algorithmic optimization already presented in \cite{Tsamardinos2006a} in order to avoid performing twice the same conditional independence test. Assuming variable $Y$ enters in $S$ at iteration $n$, so that $S_{n+1} = S_n \cup {Y}$, then the minimum association (maximum $p$~value) between any eligible variable $X$ and the target $T$  conditioned on  any subset of $S_{n+1}$ can be written as: 
\begin{align*}
    \max \left(\max_{Z \subset S_{n} \setminus X} p_{XT.Z}, \max_{Z \subset S_{n} \setminus X} p_{XT.{Z\cup Y}} \right)
\end{align*}
That means that at each iteration only the conditioning sets including the new variable $Y$ should be taken in consideration for assessing $p_{XT.Z}$, if the quantity $\max_{Z \subset S_{n} \setminus X} p_{XT.Z}$ has been previously stored.
\item Caching intermediate results: The \code{SES} function can re-use the results of conditional independence tests calculated in a previous run in order to speed-up successive computations. This feature is specifically devised for cases when the method must be run on the same data with different configuration of the hyper-parameters $a$ and $k$.
\item Parallel computing: the first step of the SES algorithm separately assesses the univariate association of each variable with the target $T$; this is a prototypical example of embarrassingly parallel task, that can be executed on multiple CPUs, by setting the \code{ncores} argument of the \code{SES} function equal to $2$ or more.
\end{itemize}

\section[Using SES]{Using \code{SES}}
In this section, we provide examples of the use of the \code{SES} function on simulated, continuous data. All examples were run with \pkg{MXM} version $0.8$.

\subsection[Installing and loading the package]{Installing and loading the package}
\pkg{MXM} and its dependencies are available from the Comprehensive \proglang{R} Archive Network (CRAN). The package does not require any external dependency for data analyses tasks that can be addressed with the \code{testIndFisher} conditional independence test (i.e., both predictors and outcome are continuous variables). A number of external packages are required for using the other conditional independence tests: \pkg{ordinal} \citep{ordinal_ref} and \pkg{nnet} \citep{nnet_ref} for the \code{testIndLogistic} with ordinal and multinomial outcome, respectively. Package \pkg{survival} is needed for the \code{censIndCR}, \code{censIndWR} and \code{testIndClogit}, while \pkg{quantreg} \citep{quantreg_ref} and \pkg{betareg} \citep{betareg_ref} are necessary for the \code{testIndRQ} and \code{testIndBeta}  respectively. Test \code{testIndZIP} is based on package \pkg{pscl} \citep{zeileis2007}. Package \pkg{MASS}  \citep{nnet_ref} is required for performing some of the log-likelihood ratio tests, for the robust version of the Fisher's test and for \code{testIndNB}. Packages \pkg{gRbase}\footnote{Some dependencies of \pkg{gRbase} package are not available on CRAN, however the users can install them directly from the Bioconductor repository.} \citep{grbase_ref} and \pkg{hash} \citep{hash_ref} are suggested for faster computations, while \pkg{foreach} \citep{Revolution2015a} and \pkg{doParallel} \citep{Revolution2015b} allow for parallel computing during the first step of the algorithm. Finally, \code{SES} supports \code{ExpressionSet} objects as input if the Bioconductor package \pkg{Biobase} is present and loaded.

\begin{CodeInput}
R> install.packages("MXM")
R> library("MXM")
\end{CodeInput}

\subsection[Discovering multiple feature signatures]{Discovering multiple feature signatures}
\label{sec:applySES}
In the following example we simulate a simple continuous data set where the target variable is associated with a subset of the predictors. Collinear variables are then included in the data set in order to create equivalent signatures. \code{SES} is then run with fixed threshold $a$ and maximum conditioning set $k$. Successively, we re-run the \code{SES} function with a different configuration on the same data, but this time we re-use the $p$~values previously computed and stored as a hash object. The results show both the capability of \code{SES} in retrieving the correct equivalent signatures and the gain in computational time ensured by the hash-based mechanism.
 
First run of \code{SES} on simulate data:

\begin{CodeInput}
R> set.seed(12345678)
R> install.packages("hash")
R> require(hash)
R> dataset <- matrix(runif(1000 * 300, 1, 100), nrow = 1000, ncol = 300)
R> target = 3 * dataset[ , 10] + 2 * dataset[ , 200] + 
+    3 * dataset[, 20] + runif(1000, 0, 10)
R> dataset[ , 15] <- dataset[ , 10]
R> dataset[ , 250] <- dataset[ , 200]
R> dataset[ , 230] <- dataset[ , 200]
R> system.time(sesObject <- SES(target, dataset, max_k = 5, 
+    threshold = 0.2, test = "testIndFisher", hash = TRUE, 
+    hashObject = NULL))
\end{CodeInput}

The output of the \code{SES} function is an object of the class \code{SESoutput} with fields:

\begin{itemize}
\item \code{selectedVars}: The selected variables, i.e., the signature of the target variable.
\item \code{selectedVarsOrder}: The order of the selected variables according to increasing $p$~values.
\item \code{queues}: A list containing lists (queues) of equivalent features, one for each variable included in \code{selectedVars}. A signature equivalent to \code{selectedVars} can be built by selecting a single feature from each queue.
\item \code{signatures}: A matrix reporting all equivalent signatures (one signature for each row).
\item \code{hashObject}: The \code{hashObject} caching the statistic calculated in the current run.
\item \code{pvalues}: This vector reports the strength of the association of each predictor with the target, in the context of all other variables. Specifically, for each variable $X$ the maximal $p$~value found over all $ind(X,T|\textbf{Z})$ executed during the algorithm is reported. Lower values indicate higher association.
\item \code{stats}: the statistics corresponding to the reported \code{pvalues}.
\item \code{max_k}: The \code{max_k} option used in the current run.
\item \code{threshold}: The threshold option used in the current run.
\item \code{runtime}: The run time of the algorithm. A numeric vector. The first element is the user time, the second element is the system time and the third element is the elapsed time.
\item \code{test}: The name of the statistical test used in the current run.
\item \code{rob}: The value of the robust option, either TRUE or FALSE.
\end{itemize}

Generic functions implemented for the \code{SESoutput} object are:
\begin{itemize}
\item \code{summary(x = \code{SESoutput})}: Summary view of the \code{SESoutput} object.
\item \code{plot(object = \code{SESoutput}, mode = "all")}: Bar plots of the $p$~values for the current \code{SESoutput} object in comparison to the threshold. Argument \code{mode} can be either "all" or "partial", using only the first 500 $p$~values of the object.
\end{itemize}

\begin{CodeInput}
R> summary(sesObject)
\end{CodeInput}
\begin{CodeOutput}
Selected Variables: [1]  10  20 200

Selected Variables ordered by pvalue: [1]  10  20 200

Queues' summary (# of equivalences for each selectedVar):

                 10 20 200
#of equivalences  2  1   3

Number of signatures: [1] 6

hashObject summary:
            Length Class Mode
stat_hash   180    hash  S4  
pvalue_hash 180    hash  S4  

Summary of the generated pvalues matrix:
   Min. 1st Qu.  Median    Mean 3rd Qu.    Max. 
 0.0000  0.3115  0.5062  0.5142  0.7223  1.0000 

Summary of the generated stats matrix:
   Min. 1st Qu.  Median    Mean 3rd Qu.    Max. 
 0.0000  0.2237  0.5300  0.5626  0.8646  1.2740 

max_k option: [1] 5

threshold option: [1] 0.2

Test: testIndFisher
Total Runtime:
   user  system elapsed 
   0.12    0.00    0.12 

Robust:
[1] FALSE
\end{CodeOutput}

Variable 20 {\it must} be included in the final model. The user can then choose one predictor between variable 10 and another, and one between variable 200 and another two. The resulting six equivalent model have approximately the same predictive performance and are all based on three predictors. 

We now re-apply the \code{SES} function \emph{on the same data} by using the cached statistics used in the previous run. The results are identical, and the computational time significantly decreases.

\begin{CodeInput}
R> hashObj <- sesObject@hashObject
R> sesObject2 <- SES(target, dataset, max_k = 2, threshold = 0.01, 
+    test = "testIndFisher", hash = TRUE, hashObject = hashObj)
R> summary(sesObject2)
\end{CodeInput}
\begin{CodeOutput}
Selected Variables: [1]  10  20 200

Selected Variables ordered by pvalue: [1]  10  20 200

Queues' summary (# of equivalences for each selectedVar):
                 10 20 200
#of equivalences  2  1   3

Number of signatures: [1] 6

hashObject summary:
            Length Class Mode
stat_hash   180    hash  S4  
pvalue_hash 180    hash  S4 

Summary of the generated pvalues matrix:
   Min. 1st Qu.  Median    Mean 3rd Qu.    Max. 
 0.0000  0.2399  0.4598  0.4780  0.6985  1.0000 

Summary of the generated stats matrix:
   Min. 1st Qu.  Median    Mean 3rd Qu.    Max. 
 0.0000  0.3815  0.7222  0.7960  1.1600  2.3590 

max_k option: [1] 2

threshold option: [1] 0.01

Total Runtime:
   user  system elapsed
   0.01    0.00    0.01
   
Robust:
[1] FALSE
\end{CodeOutput}

\subsection[Identifying the best combination of SES hyper-parameter]{Identifying the best combination of SES hyper-parameter}
Selecting the best configuration of hyper-parameters is an important step in any data analysis task; finely tuning a statistical method often allows to achieve significantly better performances than naively using the default configuration. The package \pkg{MXM} provides a dedicate function, namely \code{cv.ses}, for automatically identify the optimal configuration for the SES algorithm hyper-parameters $a$ and $k$. This function internally applies a model selection schema based on stratified cross-validation \citep{Tsamardinos2014Performance-EstimationOptimization}. 

More in detail, \code{cv.ses} partitions the available data in a given number of folds, each containing approximately the same number of samples. Each fold is used in turn as test set, while the remaining data form the training set. The latter is used for training a predictive model on the features selected by \code{SES}, model that is successively applied on the test set for obtaining testable predictions. Performances are computed for each fold and then averaged. This whole procedure is repeated for each combination of $a$ and $k$ over their respective, user-specified ranges, and the optimal configuration $\{a^*, k^*\}$ correspond to the values that produced the best average performances. The users can either provide their own pre-specified folds, or have them generated internally within \code{cv.ses} by the function \code{generateCVRuns} of the package \pkg{TunePareto} \citep{TunePareto_ref}.

The type of predictive model to fit on the training data, as well as the performance metric to use depends on the data analysis task at hand. For classification tasks, logistic regression and the receiver operator characteristic (ROC) area under the curve \citep[AUC, ][]{Fawcett2006} are the default choice. The AUC is computed with the \pkg{ROCR} \citep{ROCR_ref}. Regression problems are addressed with standard linear regression and the mean square error (MSE) metric, the latter defined as $\sum_i(y_i-\hat{y_i})^2/n$, where $n$ is the number of test instances and $y_i$, $\hat{y}_i$ are the actual target value and the prediction for instance $i$, respectively. Survival analysis tasks require specialized methods, namely the Cox proportional-hazards model \citep{Cox1972} and the concordance index \citep[CI,][]{Harrell2001RegressionAnalysis} performance metric. The CI has an interpretation similar to the AUC, ranging in $[0,1]$ with $0.5$ indicating random predictions and $1$ corresponding to a perfect rank of the test instances. Package \pkg{Hmisc} \citep{Hmisc_ref} is required for the computation of the CI metric. The user has also the possibility of providing customized functions for predictive modeling and performance evaluation. The signature of the \code{cv.ses} function is the following:

\begin{CodeInput}
R> cv.ses(target, dataset, kfolds = 10, folds = NULL, alphas = NULL, 
+    max_ks = NULL, task = NULL, metric = NULL, modeler = NULL, 
+    ses_test = NULL)
\end{CodeInput}

The argument \code{target} and \code{dataset} are as in the \code{SES} function. Other arguments are specified below.

\begin{itemize}
\item \code{kFolds}: The number of folds to partition the data in.
\item \code{folds}: A list specifying the folds to use. If provided than \code{kFolds} is ignored.
\item \code{alphas} and \code{max_ks}: The ranges of values to be evaluated for the hyper-parameters $a$ and $k$, respectively.
\item \code{task}: A character specifying the type of task to perform:``C'' for classification, ``R'' for regression and ``S'' for survival analysis.
\item \code{metric}, \code{modeler}, \code{ses_test}: user-specified functions for the performance metric, predictive modeling and conditional independence test to use, respectively.
\end{itemize}

We now apply the \code{cv.ses} function to the simulated data presented in Section \ref{sec:applySES}:

\begin{CodeInput}
R> cv.ses.object = cv.ses(target, dataset, kfolds = 10, task = "R")
\end{CodeInput}

The best SES configuration and its respective performance can be easily retrieved:

\begin{CodeInput}
R> cv.ses.object$best_performance
\end{CodeInput}
\begin{CodeOutput}
[1] -8.794793
\end{CodeOutput}
\begin{CodeInput}
R> cv.ses.object$best_configuration
\end{CodeInput}
\begin{CodeOutput}
$id
[1] 2

$a
[1] 0.1

$max_k
[1] 2
\end{CodeOutput}

\section[Experimental validation]{Experimental validation}
We further evaluate the capabilities of the SES algorithm and \pkg{MXM} package on real data. Particularly, we aim at investigating (a) if the signatures retrieved by the algorithm provide statistically-equivalent predictive performances, and (b) whether these performances are comparable with the ones provided by the state-of-the-art feature selection algorithm LASSO, as implemented in the \proglang{R} package \pkg{glmnet} \citep{glmnet_ref}. All data and scripts for replicating the results of this comparison are freely available as supplementary material.

\subsection[Data sets description]{Data sets description}
We use three different data sets for our experiments. All data sets are formed by continuous predictors, but largely differ in the number of samples / variables and in the type of outcome (see Table \ref{table:Dataset}). Moreover, each data set comes from a different application domain. The first, Breast Cancer, is targeted at the discrimination of estrogen-receptor positive (ER+) or estrogen-receptor negative (ER-) tumors using gene expression measures. This data set is publicly available in the package \pkg{breastCancerVDX} \citep{bc_ref}. The AcquaticTox data set leads to a Quantitative Structure-Activity Relationship (QSAR) regression problem. Data are freely available from the package \pkg{QSARdata} \citep{qs_ref}. The task here is to predict the toxicity of 322 different compounds on the basis of a set of 6652 molecular descriptors produced by the software DRAGON (Talete Srl, Milano Italy). The Vijver-2002 data \citep{vandeVijver2002ACancer} contains the expression measures of breast cancer patients and the aim is to relate them with their survival time.

\begin{table}[!ht]
\centering
\begin{tabular}{|p{3cm}| p{2cm} | p{2cm}| p{2.2cm}| p{3cm}|}
	\hline \rule[-2ex]{0pt}{5.5ex} Name & $\#$ samples & $\#$ variables & Task & Outcome \\
	\hline \rule[-2ex]{-2pt}{5.5ex} Breast Cancer & 17816 & 286 & Classification analysis & Binary, rarest class frequency: $36\%$\\
	\hline \rule[-2ex]{-2pt}{5.5ex} AquaticTox & 322 & 6652 & Regression analysis & Continuous \\
	\hline \rule[-2ex]{-2pt}{5.5ex} Vijver-2002 & 295 & 70 & Survival analysis & Right-censored, number of events: $88$ \\
	\hline
\end{tabular}
\caption{Data sets used for the experiments. For each data set the table reports the number of samples, the number of variables/predictors, task to accomplish (classification/regression/survival analysis) and the type of outcome. References for each data set are reported in the text.}
\label{table:Dataset}
\end{table}

\subsection[Experimentation protocol]{Experimentation protocol}
\subsubsection[Derivation and assessment of predictive models]{Derivation and assessment of predictive models}\label{sec:modelDerivation}
In order to empirically evaluate the performance of the proposed method we have repeated the following experimentation procedure $500$ times, each time using different data splits.\\
First, data are split in a training set $D_{train}$ and in a hold-out set $D_{holdout}$, each containing $50\%$ of all samples. The best hyper-parameter configuration for \code{SES} is identified on the training set through a ten-fold cross-validation model selection procedure, where the \code{SES} hyper-parameters are varied within $a \in \left[0.01, 0.05, 0.1\right]$ and $k \in \left[3, 5\right]$. \code{SES} is then run on the whole $D_{train}$ with the best hyper-parameters for identifying the optimal signatures. A predictive model for each signature is finally estimated based on $D_{train}$ and applied on $D_{holdout}$ for estimation of the performance. Logistic, linear and Cox regression procedures are used for obtaining the predictive models, depending on whether the outcome is binary, continuous or time-to-event, respectively. Appropriate performance metrics are used accordingly: AUC is used for binary classification, continuous outcomes are evaluated through the MSE, while CI is used for evaluating the performance of Cox regression models. MSE quantifies the prediction error, thus lower values indicate better performances, while the reverse holds for AUC and CI. Data splitting is stratified for classification and survival tasks, i.e., an equal proportion of instances of each class (or of censored/non-censored cases) is kept in each data split.

\subsection[Contrasting against LASSO]{Contrasting against LASSO}
SES and LASSO are used in turn as feature selection algorithm in the experimentation protocol described in Section \ref{sec:modelDerivation}, and they are compared on the basis of the performances obtained on all $D_{holdout}$. In each repetition the same data split is used for both algorithms, in order to ensure a fair comparison. Recall that LASSO selects only a single subset of relevant variables, while SES potentially retrieves multiple signatures. Thus we arbitrarily select the first signature retrieved by SES for comparison with LASSO. This is not necessarily the best one, and can be deemed to be chosen with a systematic random sampling.
In the cross-validation step of the experimentation protocol, the range of values over which we optimize the hyper-parameter $\lambda$ for the LASSO algorithm is automatically determined by the least angle regression \citep[LARS, ][]{Efron2004LeastRegression} fitting procedure.

\subsection[Results]{Results}\label{sec:res}
\subsubsection[Assessing the equivalence of SES signatures]{Assessing the equivalence of SES signatures}
Table \ref{table:claim1} reports the distribution over 500 repetitions of the number of signatures identified by SES for each dataset. Each row refers to one dataset, while each column to a given number of signatures. The results indicate that the number of retrieved signature is highly dependent upon the specific dataset. Particularly, both AquaticTox and Vijver-2002 tend to produce a large number of equivalent signatures, while a single signature is found for the Breast Cancer dataset $301$ times in $500$ repetitions. Interestingly, the number of retrieved signatures is highly variable across repetitions: for the AquaticTox dataset, simply splitting the data in different $D_{train}$ and $D_{holdout}$ sets lets the number of signatures range from $1$ to $292032$. This shows that the detection of equivalent features is strongly influenced by the specific sample at hand.

\begin{table}[!t]
\centering
\rule[-2ex]{0pt}{5.5ex}Number of signatures\\
\begin{tabular}{|p{3cm}| p{0.7cm} | p{0.7cm} | p{0.7cm} | p{0.7cm} | p{0.7cm} | p{0.7cm} | p{0.7cm} | p{0.7cm} | p{0.7cm} | p{0.7cm}|}
	\hline \rule[-2ex]{0pt}{5.5ex}  & 1 & 2 & 3 & 4 & 5 & 6 & 7 & 8 & 9 & 10+ \\
	\hline \rule[-2ex]{-2pt}{5.5ex} Breast Cancer & 301 & 136 & 7 & 40 & 1 & 2 & 0 & 3 & 1 & 9 \\
	\hline \rule[-2ex]{-2pt}{5.5ex} AquaticTox & 17 & 17 & 15 & 12 & 9 & 16 & 4 & 18 & 13 & 379 \\
	\hline \rule[-2ex]{-2pt}{5.5ex} Vijver-2002 & 31 & 181 & 13 & 94 & 3 & 55 & 1 & 37 & 3 & 82 \\
	\hline
\end{tabular}
\caption{Frequency of signature multiplicity. Each cell reports how many times $j$ equivalent signatures are retrieved for its corresponding data set (row), where $j$ is the number reported on top of the cell's column. The notation $10+$ indicates $10$ or more signatures.}
\label{table:claim1}
\end{table}

\begin{figure}[!tb]
    \centering
    \includegraphics[width=0.8\textwidth]{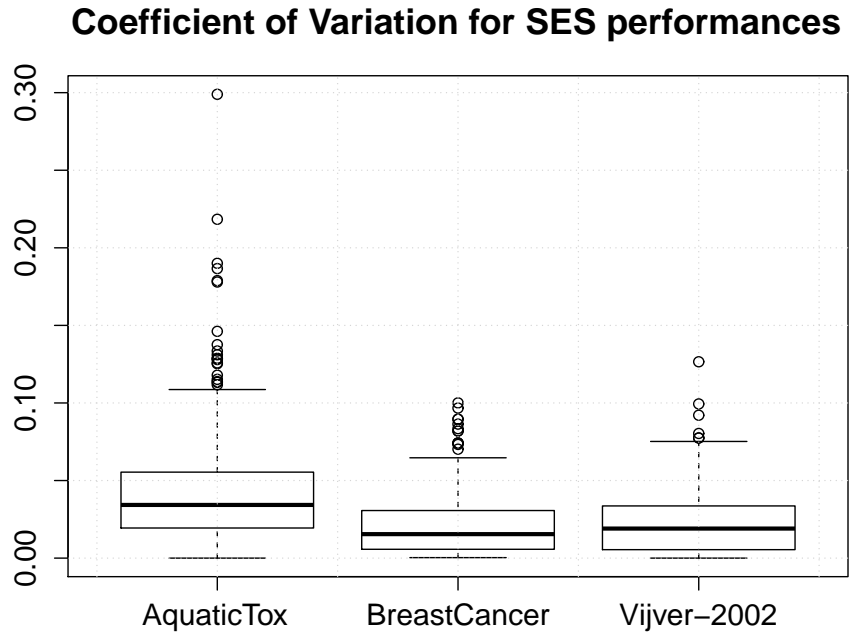}
    \caption{Boxplot of the SES performances' coefficient of variation across the 500 iterations for each dataset.}
    \label{fig:boxplot_coefvar}
\end{figure}

We now investigate whether the retrieved signatures achieve performances that are actually equivalent. For each data set and for each repetition where at least two signatures are retrieved, we compute the SES performances' coefficient of variation (CV). The CV is defined as the ratio between standard deviation and mean value, and it measures the dispersion of a distribution standardized with respect to the magnitude of its measurements. Figure \ref{fig:boxplot_coefvar} and Table \ref{table:claim2CV} show that in all data sets the median CV value is well below $5\%$, indicating that within each repetition the performances of the retrieved signatures are extremely close to each other. The AquaticTox data set produces the highest CV values, marked as circles/outliers in Figure \ref{fig:boxplot_coefvar} (two extreme CV values, reaching $1.19$ and $0.97$ respectively, were removed for the sake of figure readability). We also observe that the higher the number of signatures, the higher the coefficient of variation (Spearman correlation: $0.69$ Vijver, $0.25$ Breast Cancer, $0.45$ AquaticTox data set, $p$~value $< 0.001$ in all cases). This result is not unexpected. When few signatures are retrieved, each signature differs from the other for only one or two features, and thus their predictive performances are expected to be similar. When thousands of signatures are produced, their heterogeneity increases, as well as the deviation of their performances. It can be concluded though that the algorithm is generally stable, with very rare exceptions, and leads in general to signatures with very close predictive performance. It could be argued that the variation in the estimated predictive performance is often an order of magnitude lower than the performance estimates themselves. 

\begin{table}[!b]
\centering
\begin{tabular}{|p{3cm}| p{3cm} | p{3cm} | p{3cm}|}
	\hline \rule[-2ex]{0pt}{5.5ex}  & $2.5\%$ & Median & $97.5\%$ \\
	\hline \rule[-2ex]{-2pt}{5.5ex} Breast Cancer & 0.06$\%$ & 2.17$\%$ & 8.36$\%$ \\
	\hline \rule[-2ex]{-2pt}{5.5ex} AquaticTox & 0.06$\%$ & 4.20$\%$ & 12.80$\%$ \\
	\hline \rule[-2ex]{-2pt}{5.5ex} Vijver-2002 & 0.04$\%$ & 2.18$\%$ & 6.44$\%$ \\
	\hline
\end{tabular}
\caption{Quantiles of the coefficient of variation (CV) of the SES performances. Results are reported separately for each data set (rows).}
\label{table:claim2CV}
\end{table}

\subsubsection[Contrasting SES and LASSO]{Contrasting SES and LASSO}
\begin{figure}[!t]
    \centering
    \includegraphics[width=0.8\textwidth]{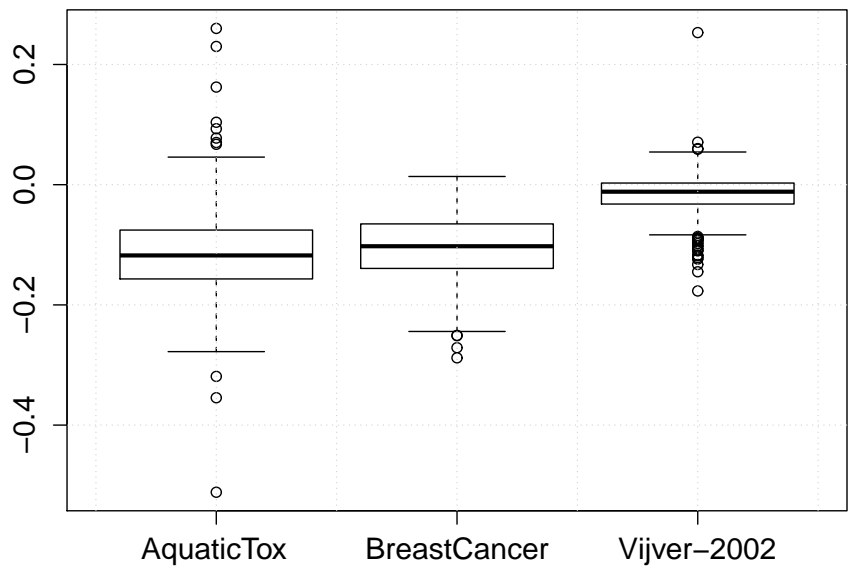}
    \caption{Boxplot of the difference among SES and LASSO performances across the 500 iterations for each dataset.}
    \label{fig:boxplot_ses-lasso}
\end{figure}

Table \ref{table:claim2std} shows the $95\%$ confidence intervals of the paired differences in performance between SES and LASSO, computed over $500$ repetitions. For each data set, the differences are computed in such a way that positive values indicate SES outperforming LASSO, and vice versa. The table shows that for both the Vijver-2002 and AquaticTox data sets the confidence intervals cross zero, thus on these two data sets the SES and LASSO methods are not statistically different at $0.05$ significance level. LASSO performs slightly better in the Breast Cancer data set though. Figure \ref{fig:boxplot_ses-lasso} reports the distribution for the differences in performances between the two methods. Here the equivalence between the two methods on the Vijver-2002 is even more evident, while differences in performances for the AquaticTox dataset show quite a large variability.\\
Table \ref{table:claim3NumVars} shows the distribution of the number of selected variables over the $500$ repetitions. SES is generally quite  parsimonious, and it usually selects the same number of variables, independently by the data split, as demonstrated by the low standard deviations over the $500$ repetitions. In contrast, the number of variables selected by LASSO widely varies across repetitions. SES also selects much fewer variables than LASSO for both the AquaticTox and Vijver-2002 data sets, while for the Breast Cancer dataset LASSO produces only slightly more parsimonious models but again with larger variability.

\begin{table}[!t]
\centering
\begin{tabular}{|p{3cm}| p{3cm} | p{3cm} | p{3cm}|}
	\hline \rule[-2ex]{0pt}{5.5ex}  & $2.5\%$ & Mean & $97.5\%$ \\
	\hline \rule[-2ex]{-2pt}{5.5ex} Breast Cancer & -0.222833 & -0.104123 & -0.001349 \\
	\hline \rule[-2ex]{-2pt}{5.5ex} AquaticTox & -0.228774 & -0.115798 & 0.027270 \\
	\hline \rule[-2ex]{-2pt}{5.5ex} Vijver-2002 & -0.096756 & -0.016740 & 0.033616 \\
	\hline
\end{tabular}
\caption{Quantiles of the difference in performance between SES and LASSO. Positive values indicate SES outperforming LASSO.}
\label{table:claim2std}
\end{table}

\section[Discussion and conclusions]{Discussion and conclusions}
In the present work we introduced the \proglang{R} package \pkg{MXM}, which implements the SES algorithm for selecting statistically-equivalent sets of predictive signatures. The package further implements a number of conditional independence tests able to cope with a variety of data types. These tests can be used alone (for inference or causal discovery) or in conjunction with the \code{SES} function, in order to deal with several data-analysis tasks, including (multi-class) classification, regression and survival analysis.\\
We used three real-world, publicly available data sets from different application areas for evaluating the capabilities of the software. Multiple signatures were actually identified for all data sets, indicating that equivalences among predictors are frequently present in real applications. Deviation among the signatures' performances proved to be particularly low, indicating that the signatures have almost equal predictive power.\\
We further contrasted the performance of the SES algorithm against the LASSO method. We attempted to have a comparison as fair as possible, so we always compared the LASSO signature against the first one retrieved by SES. In the context of our experiments, SES was more stable in terms of number of variables selected across different data splits, while LASSO in general selects a higher number of variables. The two methods had quite comparable performance, with LASSO performing slightly better in the Breast Cancer example. These results are in agreement with previous theoretical results \citep{Meinshausen2006High-dimensionalLasso}: under some quite general conditions, LASSO retrieves a super set of the Markov-Blanket of the target variable, meaning that all variables needed for optimal prediction plus some noisy variables are selected. In contrast, SES is devised for retrieving the Parent-Children set of the target variable, i.e., a subset of the Markov-Blanket. Thus, it is not surprising that in our experimentation SES selects fewer variables and does not outperform LASSO. We also note that these results may be influenced by the restricted range of values over which SES hyper-parameters $a$ and $k$ have been optimized.\\
The aim of this paper is not an assessment of SES, of course;  
and results in Section \ref{sec:res} are not conclusive. A more extensive comparison study is currently under preparation in order to exhaustively evaluate SES capabilities and contrast its performance against a range of feature selection methods.

\begin{table}[!t]
\centering
\begin{tabular}{|p{3cm}| p{2cm}| p{2.5cm} | p{2cm} | p{2cm} |}
	\hline \rule[-2ex]{0pt}{5.5ex}  & Average SES & Average LASSO & StD. SES & StD. LASSO \\
	\hline \rule[-2ex]{-2pt}{5.5ex} Breast Cancer & 13.29 & 10.62 & 5.91 & 15.43 \\
	\hline \rule[-2ex]{-2pt}{5.5ex} AquaticTox & 5.68 & 160.75 & 1.80 & 66.34 \\
	\hline \rule[-2ex]{-2pt}{5.5ex} Vijver-2002 & 3.24 & 10.50 & 0.98 & 3.64 \\
	\hline
\end{tabular}
\caption{Distribution of the number of variables selected by SES and LASSO. For each method and data set both the average number and the standard deviation (St.D.) of selected variables is reported.}
\label{table:claim3NumVars}
\end{table}

In conclusion, our limited experiments indicate that:
\begin{itemize}
\item Multiple, equally-performing signatures naturally occur in real-world data sets, either due to equivalence among predictors or to impossibility to distinguish them due to limited sample size. In either case, this phenomenon should be duly taken into account while retrieving predictive feature subsets.
\item The signatures retrieved by the SES algorithm provide predictive performances extremely close to each other in all data sets included in our analyses, demonstrating in fact to be equally-predictive.
\item SES and LASSO provide comparable results, and SES is generally more parsimonious and sheds light on the characteristics of the problem at hand by identifying equivalences hidden into the data.
\end{itemize}
We keep developing \pkg{MXM} by adding new conditional independence tests, as well as new functionalities. For example. the MMPC algorithm, which performs feature selection without providing multiple solutions, and the PC and MMHC algorithms, two methods for constructing the skeleton of a Bayesian network. 
Future work will focus on both algorithmic and implementation improvements. In future extensions SES will attempt to retrieve the Markov Blanket of the target variable, i.e., the variables set with theoretically the highest predictive power. The aggregation of models trained on equivalent signatures for improving predictive performances is also under consideration. In addition, we aim at extending \pkg{MXM} in the areas of model selection and performance estimation \citep{Tsamardinos2014Performance-EstimationOptimization}, two fields closely related to the problem of feature selection.

\section{Acknowledgments}
The work was co-funded by the STATegra EU FP7 project, No 306000, by the EPILOGEAS GSRT ARISTEIA II project, No 3446, and by the European Research Council (ERC) project No 617393, "CAUSALPATH - Next Generation Causal Analysis". We sincerely thank Damjan Krstajic and Giorgos Borboudakis for their invaluable comments, suggestions and critical reading of the manuscript.

\bibliographystyle{jss}
\bibliography{athineou}
%

\end{document}